\documentclass[a4paper]{article}
\usepackage{INTERSPEECH2021}
\usepackage{amsmath}
\usepackage{amssymb}
\usepackage{bm}

\title{Pre-training for Spoken Language Understanding with Joint Textual and Phonetic Representation Learning}
\name{Qian Chen, Wen Wang, Qinglin Zhang}
\address{Speech Lab, Alibaba Group}
\email{\{tanqing.cq, w.wang, qinglin.zql\}@alibaba-inc.com}

\begin{document}

\maketitle
\begin{abstract}
In the traditional cascading architecture for spoken language understanding (SLU), it has been observed that automatic speech recognition errors could be detrimental to the performance of natural language understanding. End-to-end (E2E) SLU models have been proposed to directly map speech input to desired semantic frame with a single model, hence mitigating ASR error propagation. Recently, pre-training technologies have been explored for these E2E models. In this paper, we propose a novel joint textual-phonetic pre-training approach for learning spoken language representations, aiming at exploring the full potentials of phonetic information to improve SLU robustness to ASR errors. We explore phoneme labels as high-level speech features, and design and compare pre-training tasks based on conditional masked language model objectives and inter-sentence relation objectives. We also investigate the efficacy of combining textual and phonetic information during fine-tuning. Experimental results on spoken language understanding benchmarks, Fluent Speech Commands and SNIPS, show that the proposed approach significantly outperforms strong baseline models and improves robustness of spoken language understanding to ASR errors.

\end{abstract}
\noindent\textbf{Index Terms}: spoken language understanding, pre-training, joint text and speech representation learning

\section{Introduction}
\label{sec:introduction}
Spoken language understanding (SLU) is a critical component for goal-oriented spoken dialogue systems which facilitate various voice assistants. SLU interprets a spoken query by predicting the intent of the query (intent classification, IC) and predicting the semantic concepts (slots) (slot filling, SF), respectively. For example, the intent for a speech command ``play a popular song by brian epstein'' to a voice assistant is \emph{PlayMusic}, and the slots are \emph{sort:popular, music\_item:song, artist:brian epstein}.  
The conventional architecture for SLU is a cascaded paradigm, where an automatic speech recognition (ASR) system converts speech signals to text and then natural language understanding (NLU) systems predict intent and  slots on the ASR output. It has been observed that ASR errors could cause severe performance degradation on the downstream NLU systems~\cite{DBLP:conf/flairs/WangALT11}. To improve SLU robustness to ASR errors, many prior works explore multiple ASR hypotheses. \cite{MorbiniSLT2012} developed a reranking approach combining ASR N-bests and NLU. \cite{DBLP:journals/csl/Hakkani-TurBRT06} and \cite{Tur13semanticparsing} exploited word confusion networks (WCN) for call classification and WCN-based conditional random fields (CRF) for SF. \cite{DBLP:conf/interspeech/ShivakumarYG19} used unsupervised word representations incorporating acoustic relationships learned from WCNs for IC. \cite{DBLP:conf/interspeech/LadhakGDMRH16} used lattice embeddings computed from RNN for IC. \cite{DBLP:conf/lrec/SimonnetGCE18} augmented the SLU training manual transcripts by simulating ASR errors on them, where the ASR confusability functions for error simulation are learned from the specific ASR system in cascaded SLU. Recently, end-to-end approaches have been proposed to address error propagation and to enable joint optimization of ASR and NLU. E2E approaches directly map speech to desired semantic frame with a single model. Pre-training text or speech representations has been introduced to alleviate data scarcity, including encoding speech and text separately (two-stream model) and jointly (single-stream model). Some E2E SLU work models IC only \cite{DBLP:conf/icassp/Serdyuk18,DBLP:conf/interspeech/LugoschRITB19,DBLP:conf/icassp/Sari0H20,DBLP:conf/icassp/WangWCXN20,DBLP:conf/interspeech/ChoKYK20,DBLP:conf/interspeech/radfar20,DBLP:journals/corr/abs-2102-07370} while some works jointly perform IC and SF and optionally generate ASR transcripts~\cite{DBLP:conf/slt/HaghaniNBCGMPQW18,DBLP:conf/slsp/TomashenkoCELM19,DBLP:conf/interspeech/RaoRDBR20,DBLP:journals/corr/abs-2102-06283}. 
Different from the past E2E approaches, this work aims at improving NLU performance on ASR 1-best in the conventional cascaded architecture, considering situations when downstream NLU systems have access to only ASR 1-best as input, without access to the original audio.

Phonetic information has also been explored to improve robustness to ASR errors, but only with limited effort, such as augmenting word embeddings with phone embeddings~\cite{DBLP:conf/acl/LiuMHXH19,DBLP:journals/corr/abs-1811-00728}, using phone boundaries to compress speech features~\cite{DBLP:conf/acl/SaleskySB19}, and combining speech feature vectors and phone embeddings~\cite{DBLP:conf/acl/SaleskyB20}. In contrast, our work attempts to explore the full potential of using phonetic information to improve SLU robustness to ASR errors, by learning joint textual-phonetic representations through pre-training and also exploring phonetic information in fine-tuning.  The major contribution of this paper is two-fold: 
\begin{itemize}
\setlength\itemsep{0em}
    \item We propose a single-stream pre-trained model to learn joint textual-phonetic semantic representations for SLU. We design and study a variety of pre-training tasks for this purpose. We also investigate incorporating phonetic features in fine-tuning and the combinatory effect with the proposed pre-trained models.
    \item On SLU Fluent Speech Commands (FSC) and SNIPS benchmarks, the proposed approach consistently improves SLU performance on ASR 1-best and significantly outperforms strong baseline models.
\end{itemize}

\section{Joint Textual-Phonetic Representation Learning}
\label{sec:pretraining}
In this section, we first introduce the proposed pre-training approach for learning joint semantic representations from textual and phonetic information. We then describe the approach of combining textual and phonetic information during fine-tuning.

\subsection{Pre-trained Model} 
\label{subsec:model}
\begin{figure}[ht]
\centering
\includegraphics[width=0.48\textwidth]{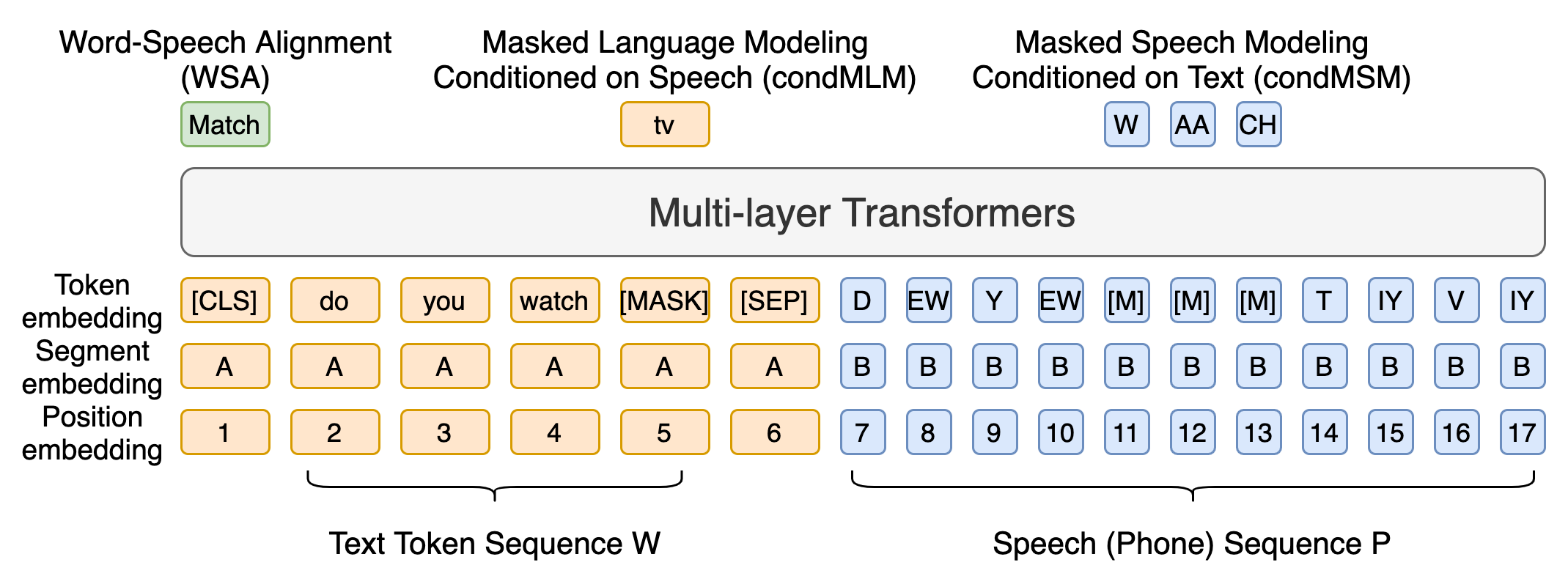}
\caption{A diagram illustrating the proposed single-stream pre-training model and pre-training tasks for joint textual and phonetic representation learning.}
\label{fig:modelarchitecture}
\end{figure}

Figure~\ref{fig:modelarchitecture} illustrates the architecture of the proposed pre-training approach. In this study, we use the manual transcripts of the Librispeech 960hrs data~\cite{librispeech} and the Fisher corpus\footnote{Fisher English Training Speech Part 1 LDC2004S13 and LDC2004T19; Part2 LDC2005S13 and LDC2005T19} as the pre-training data set (denoted $\mathcal{D}_{pt}$) for our proposed pre-trained model. We use phoneme labels to represent high-level speech features. Given each sentence $W=w_1,w_2,\ldots,w_m, W \in \mathcal{D}_{pt}$ (where $w_i$ denotes the $i$th word), we construct its phone label sequence $P=p_1,p_2,\ldots, p_n$ by looking up the phoneme sequence for each word $w_i$ in a pronunciation dictionary. We use the CMU pronunciation dictionary\footnote{http://www.speech.cs.cmu.edu/cgi-bin/cmudict}.  For words that are not covered by the CMU dictionary, we use a special token $<$UNK$>$ to represent its phone sequence. 1.3\% words in $\mathcal{D}_{pt}$ receive a $<$UNK$>$ label. The phone sequence for words with missing pronunciations can be generated with a grapheme-to-phoneme (g2p) model. 
In future work, we plan to explore phone labels generated through forced alignment~\cite{DBLP:conf/interspeech/LugoschRITB19}.

Each pair $<$W,P$>$ is one input training sample to the pre-trained model. The input is embedded through the input representation layer as the element-wise sum of token embedding, position embedding, and segment embedding. These embeddings are then fed to a multi-layer bidirectional transformer encoder to learn the joint contextualized representations for the textual and phonetic sequences. 

\subsection{Pre-training Tasks}
\label{subsec:tasks}
We design three pre-training tasks for learning joint textual-phonetic semantic representations:\emph{masked language modeling conditioned on the phone sequence (condMLM)}, \emph{masked speech modeling conditioned on the word sequence (condMSM)}, and \emph{word-speech alignment (WSA)}, inspired by UNITER~\cite{DBLP:conf/eccv/ChenLYK0G0020} which is a single-stream image-text representation model.
\newline\\
\noindent\textbf{Masked language modeling conditioned on the phone sequence.} We randomly mask $M\%$ tokens in the text sequence $W$ and replace the masked tokens with a special token $[$MASK$]$. Same as BERT,  $M\%$ masking is carried out as 80\% $[$MASK$]$ substitution and 10\% random word substitution and 10\% unchanged. The model learns to use the remaining unmasked tokens in $W$ and the entire unmasked phone sequence $P$ to predict the masked tokens, and the goal is to minimize the cross-entropy loss, denoted $\mathcal{L}_{condMLM}$.
\newline\\
\noindent\textbf{Masked speech modeling conditioned on the word sequence.} We randomly mask $N\%$ tokens in the phone sequence $P$ and replace them with $[$MASK$]$. The model then uses the remaining unmasked phones and the entire unmasked sentence $W$ to predict the masked phones. The optimization is conducted by minimizing the cross-entropy loss of this prediction, denoted $\mathcal{L}_{condMSM}$. Previous studies show whole-word masking outperforms WordPiece based masking. In addition, to avoid introducing noise from aligning the phone sequence of a word to its WordPiece tokenization, we use whole-word masking for both words and phones. That is, we mask all tokens corresponding to a word at once and similarly, all phones corresponding to a word at once. Considering the asymmetric complexity of predicting words based on phones and vice versa, we investigate \emph{oneMod} and \emph{twoMod} masking strategies. In the oneMod strategy, for each input sample, we randomly choose masking the word sequence or the phone sequence, then conduct random masking on the chosen modality, but not masking both word and phone sequences. In contrast, in the twoMod strategy, both word sequence and phone sequence are randomly masked. 
\newline\\
\noindent\textbf{Word-speech alignment.} We design a binary classification task to learn word-speech alignment between a text sequence $W$ and a phone sequence $P$. Given $<$W,P$>$ of sentences in $\mathcal{D}_{pt}$ and their phone sequences, we construct ``[CLS] W [SEP] P'' as positive samples and ``[CLS] W [SEP] P$_{rand}$'' as negative samples, where P$_{rand}$ is randomly sampled from $<$W',P'$>$ (W' $\ne$ W). The hidden state for [CLS] is fed into a softmax classifier to decide whether the word sequence and phone sequence match or not. The training objective is minimizing the softmax cross-entropy loss, denoted $\mathcal{L}_{WSA}$.

The overall training objective is the combination of the objectives of these pre-training tasks. For each positive sample of the WSA task, loss is $\mathcal{L}_{condMLM} + \mathcal{L}_{condMSM} + \mathcal{L}_{WSA}$. For each negative sample of the WSA task, condMLM and condMSM fall back to MLM and MSM, respectively; and loss is $\mathcal{L}_{MLM} + \mathcal{L}_{MSM} + \mathcal{L}_{WSA}$.

\label{sec:model}

\subsection{Fine-tuning for SLU}
\label{subsec:finetuning}
Fine-tuning for SLU in this work includes intent classification(IC) only,  and jointly performing IC and slot filling (SF) in a multi-task learning framework~\cite{DBLP:journals/corr/abs-1902-10909}. We prepend the special token [CLS] to each tokenized sequence in the SLU training set and append [SEP] to it. Given this input sequence $x_1,\cdots,x_T$ to a pre-trained model, the output hidden states are denoted $h_1,\cdots,h_T$. IC is then modeled as:
\begin{align}
\label{equ:intent}
p^I = \mathrm{softmax}({W}^I F^I({h}_{[CLS]}) + {b}^I) \end{align}
\noindent where $F^I$ is a non-linear feed-forward layer with tanh activation. During inference, the intent label is predicted by $\mathrm{argmax}(p^I)$. 
During IC-only finetuning, the model is trained via minimizing the softmax cross-entropy loss of IC.

For jointly performing IC and SF, in addition to modeling IC as Eq.~\ref{equ:intent}, the final hidden states of other tokens, that is, $[h]_2^T$, are fed into a softmax layer to classify over the SF labels in the BIO scheme. Facilitating compatibility with the WordPiece tokenization, each tokenized input word is processed by a WordPiece tokenizer and the hidden state corresponding to the first sub-token is used as input to the softmax classifier.
\begin{align}
p^S_i = \mathrm{softmax}({W}^S {h}_i + {b}^S)
\end{align}
\noindent where \({h}_i\) is the hidden state corresponding to the first sub-token of $x_i$. The joint model is fine-tuned via minimizing the sum of the softmax cross-entropy losses of IC and SF.

We propose an approach to introduce phone labels during SLU fine-tuning. For the input embedding layer, given a word $w_i$ and the phone sequence $p_i^1,\ldots,p_i^l$ representing its pronunciation, the augmented input embedding for $w_i$ is computed in Eq.~\ref{eq:phoneemb}, as the weighted element-wise sum of the standard input embedding and phone embeddings\footnote{We compared computing the phone embeddings as the sum or the mean of single phone embeddings $E[p_i^j],j=1,\cdots,l$ and observed better SLU performance from sum-pooling.}, 
\begin{equation}
\label{eq:phoneemb}
E[w_i,p_i^1,\ldots,p_i^l] = E[w_i] + \beta*\sum_{j=1}^{l}E[p_i^j]
\end{equation} 
\noindent where $E(*)$ denotes trainable embedding; $E[w_i]$ denotes the standard element-wise sum of token embedding, position embedding, and segment embedding; $\beta$ is a hyperparameter optimized on the validation set of a SLU task.

\begin{table}[htb]
\renewcommand{\arraystretch}{0.9}
\begin{center}
\begin{tabular}{l l l}
\hline
\textbf{} & FSC & Snips \\
\hline
Train & 23,132 & 13,084 \\
Valid & 3,118 & 700 \\
Test & 3,793 & 700 \\
Intents & 31 & 7 \\
Slot Types & - & 39 \\
WER(\%) (valid/test) & Sys1 39.0/19.2 & 40.8/42.3~\cite{DBLP:conf/icassp/HuangC20,DBLP:conf/acl/HuangC20} \\
& Sys2 36.7/15.2 & - \\
\hline
\end{tabular}
\end{center}
\caption{Data statistics. Sys1 and Sys2 denote the two off-the-shelf Kaldi ASR systems used for decoding (Section~\ref{subsec:setup}). All Word Error Rates (WERs) reported are 1-best WER.}
\label{tab:data-statistics}
\end{table}

\begin{table*}[htb]
\renewcommand{\arraystretch}{0.9}
\begin{center}
\scalebox{0.9}{
\begin{tabular}{l c c c c c c c c}
\hline
\textbf{Model} & 
\multicolumn{4}{c}{\textbf{FSC ICAcc}} & \multicolumn{4}{c}{\textbf{Snips}} \\
& \multicolumn{2}{c}{\textbf{WER 19.2}} & \multicolumn{2}{c}{\textbf{WER 15.2}}
& \multicolumn{2}{c}{\textbf{w/o PE}} & \multicolumn{2}{c}{\textbf{w/ PE}} \\
& \textbf{w/o PE} & \textbf{w/ PE} & \textbf{w/o PE}  & \textbf{w/ PE} 
& \textbf{IcAcc} & \textbf{semER$\downarrow$} & \textbf{ICAcc} & \textbf{semER$\downarrow$} \\
\\
\hline
BERT-Base & 87.6 & 89.0 & 92.4 & 94.4 
& 82.1 & 57.1 & 82.7 & 54.6 \\
\hline
\hline
+MLM 15\%   & 88.5 & 89.1 & 93.0 & 93.5 
& 82.0 & 57.2 & 84.4 &  54.9 \\
+MLM 15\%+NSP  & 88.2 & 89.5 & 92.1 & 93.8  
& 81.1 & 57.3 & 85.0 & 55.5 \\
+condMLM 100\%+condMSM 100\%(oneMod) & \textbf{89.2} & 89.7 & \textbf{94.2} & 94.9 & \textbf{83.7} & \textbf{55.3} & \textbf{85.4} & \textbf{53.4} \\
+condMLM 30\%+condMSM 30\%(twoMod) & \textbf{89.2} & \textbf{90.5} & 93.7 & \textbf{95.1} & 82.7 & 55.8 & 83.1 & 56.0 \\
+condMLM 30\%+condMSM 30\%(twoMod)+WSA & 88.5 & 90.0 & 93.1 & 94.8 & 81.6 & 56.9 & 85.0 & 54.7 \\
+condMLM 100\%+MLM 15\%(oneMod) & 88.0 & 90.1 & 92.8 & 94.7   
& 81.7 & 56.5 & 80.9 & 56.1 \\
+condMSM 100\%+MLM 15\%(oneMod) & 88.0 & 89.9 & 92.6 & 94.9   
& 80.0 & 57.2 & 82.3 & 55.3 \\
+condMLM 100\%+condMSM 100\%+MLM 15\%(oneMod) & 88.1 & 89.4 & 93.3 & 94.4 & 81.3 & 56.7 & 84.1 & 55.7 \\
\hline
\hline
Oracle BERT-base & & & & & 93.9 & - & 94.3 & - \\
\hline
\end{tabular}
}
\end{center}
\caption{Intent classification accuracy (ICAcc) on ASR 1-best of the FSC test set with different WERs, and ICAcc and semER on ASR 1-best of the Snips test set. \textbf{w/o PE} and \textbf{w/ PE} denote fine-tuning without or with adding phone embeddings (Section~\ref{subsec:finetuning}). \textbf{x\%} indicates the masking percentage. \textbf{OneMod} denotes masking a randomly chosen single modality for each pre-training sample; whereas \textbf{twoMod} denotes masking both modalities for each sample. Except \textbf{Oracle} which conducts SLU fine-tune using the SLU train set ASR 1-best, all experiments conduct SLU fine-tuning on the SLU train set manual transcripts. All the gains from the bold-faced numbers over their respective baselines are statistically significant with $\mathbf{p<0.01}$.}
\label{tab:all-results}
\end{table*}

\begin{table}[htb]
\renewcommand{\arraystretch}{0.9}
\begin{center}
\begin{tabular}{l c}
\hline
\textbf{Pre-trained Models} & MRR \\
\hline
BERT-Base & 0.1012 \\
\hline
+MLM 15\%+NSP & 0.1180 \\
+condMLM 100\%+condMSM 100\%(oneMod) & \textbf{0.1591} \\
+condMLM 30\%+condMSM 30\%(twoMod) & 0.1396 \\
\hline
\end{tabular}
\end{center}
\caption{Mean reciprocal rank (MRR) computed on the most frequent 20 confusion word pairs in the FSC valid set ASR 1-best (WER 36.7\%), using the trained input embeddings from the pre-trained models.}
\label{tab:MRR-results}
\end{table}

\section{Experiments}
\label{sec:expts}
\subsection{Experimental Setup}
\label{subsec:setup}
We evaluate our proposed approach on two SLU benchmarks: Fluent Speech Commands (FSC)~\cite{DBLP:conf/interspeech/LugoschRITB19} and Snips~\cite{DBLP:journals/corr/abs-1805-10190}. FSC includes recordings of 248 unique English command phrases (e.g., ``turn the lights off in the kitchen'') to a virtual assistant from 77 speakers. Following prior work~\cite{DBLP:conf/interspeech/LugoschRITB19}, the three slot values annotated for each audio file (``action'', ``object'', ``location'') are combined as the intent of the utterance and we conduct IC on FSC. 
We decode 1-best for the validation and test sets of FSC with two off-the-shelf Kaldi ASR systems\footnote{Sys1: ASpIRE Chain Model: https://kaldi-asr.org/models/m1 and Sys2: Librispeech ASR Model https://kaldi-asr.org/models/m13 with trigram decoding and RNNLM rescoring. Both ASR systems use the CMU dictionary as used for phone label lookup.}. 

The second dataset is Snips~\cite{DBLP:journals/corr/abs-1805-10190}, collected from the Snips personal voice assistant. 
Since the original Snips release only comprises of text data without natural speech released, we use the ASR hypotheses for the Snips validation and test sets from~\cite{DBLP:conf/icassp/HuangC20,DBLP:conf/acl/HuangC20}\footnote{The authors synthesized audio from text using the Google TTS system and decoded with Kaldi Sys1.}. Table~\ref{tab:data-statistics} summarizes the statistics of the datasets.

\noindent\textbf{Evaluation Metrics.} We report intent classification accuracy (ICAcc) on the FSC test set ASR 1-best and both ICAcc and semantic error rate (semER)~\cite{DBLP:conf/interspeech/RaoRDBR20} on the Snips test set ASR 1-best. SemER jointly evaluates IC and SF. We count correct slots (slot names and values correctly identified), deletion errors (slot names appear in reference but not in hypothesis), insertion errors (extraneous slot names in hypothesis), and substitution errors (correct slot names in hypothesis but incorrect slot values, and IC errors). SemER is then computed as:
\begin{equation}
SemER = \underbrace{\frac{\#Del+\#Ins+\#Sub}{\#Cor+\#Del+\#Sub}}_\text{\# Slots in Reference}
\end{equation}

For the baseline pre-trained model, we use English uncased BERT-Base\footnote{https://github.com/google-research/bert}, pre-trained on the BooksCorpus~\cite{DBLP:conf/iccv/ZhuKZSUTF15} and English Wikipedia.
We further pre-train BERT on different combinations of condMLM, condMSM, and WSA tasks, as well as masked language modeling (MLM) and next sentence prediction (NSP) tasks used in BERT pre-training. We then conduct SLU fine-tuning on the pre-trained models for evaluation. We also investigate efficacy of oneMod and twoMod masking strategies and different masking percentages for condMLM and condMSM. The maximum sequence length is 256, the batch size is 64, and the number of training steps is 100K. Adam~\cite{DBLP:journals/corr/KingmaB14} is used for optimization. The initial learning rate is optimized among \{1e-4, 5e-5\} for pre-training and among \{3e-5,  5e-5\} for SLU fine-tuning. The dropout probability is 0.1. 

\subsection{Results and Analysis}
\label{subsec:results}
\noindent\textbf{SLU Results.} Table~\ref{tab:all-results} shows intent classification accuracy (ICAcc) on two sets of ASR 1-best of the FSC test set, with WERs of 19.2 and 15.2, respectively. The baseline ICAccs from fine-tuning BERT-Base are 87.6 and 92.4, comparable to 90.11 ICAcc reported in ~\cite{DBLP:conf/icassp/WangWCXN20} by pipelining an E2E ASR system unadapted to SLU datasets and BERT-Base NLU.
We further pre-train BERT on our pre-training data $\mathcal{D}_{pt}$ using MLM with 15\% masking,  as well as adding NSP. Since BERT is mostly pre-trained on written text, further pre-training BERT on $\mathcal{D}_{pt}$ composed of speech transcripts is an adaptation to reduce mismatch between written text and spoken language in the SLU tasks. Our results confirm this hypothesis as ICAccs on the two sets of ASR 1-best improve from 87.6 to 88.5, and 92.4 to 93.0. Adding NSP to MLM pre-training does not yield improvement.

For our proposed pre-training model, we initialize with BERT-Base and then further pre-train the model by combining different pre-training tasks. For example, \emph{condMLM 100\% + condMSM 100\% (oneMod)} denotes the configuration in which for each training sample, we first randomly choose the modality of speech or text to mask, then mask all tokens on the chosen modality and use the entire sequence of the other modality to recover the masked tokens. We compare SLU results on the validation set of each SLU task, from pre-training with different task combinations. We observe that the proposed model achieves significant improvement over the baseline on the validation set and also observe that the same performance rankings of pre-training configurations are retained on the SLU test set. As shown in Table~\ref{tab:all-results}, both \emph{condMLM 100\% + condMSM 100\% (oneMod)} and \emph{condMLM 30\% + condMSM 30\% (twoMod)} achieve the best ICAcc 89.2 on ASR 1-best with WER 19.2 after fine-tuning without adding phone embeddings, 1.6\% absolute gain over the baseline. Adding phone embeddings into fine-tuning achieves additional gain 1.3\%, overall \textbf{2.9\%} absolute gain (87.6 to 90.5) over baseline. On the test set with WER 15.2, similarly, pre-training with \emph{condMLM 100\% + condMSM 100\% (oneMod)} obtains 1.8\% absolute gain and adding phone embeddings into fine-tuning further improves 0.7\% absolutely and raises the overall absolute improvement to 2.5\% and the best gain is \textbf{2.7\%} absolute (92.4 to 95.1). These results demonstrate that the proposed pre-training and fine-tuning approaches significantly and consistently improve SLU performance on ASR hypotheses with different WERs.

Table~\ref{tab:all-results} also shows ICAcc and semER results on ASR 1-best of the Snips test set (WER 42.3).  The baseline ICAcc and semER from fine-tuning BERT-Base are 82.1 and 57.1. Pre-training with \emph{condMLM 100\% + condMSM 100\% (oneMod)} achieves ICAcc 83.7 and semER 55.3, after fine-tuning without adding phone embeddings, \textbf{1.6\%} and \textbf{1.8\%} absolute gains over baseline. Adding phone embedding into fine-tuning achieves additional absolute gains of 1.7\% and 1.9\%, overall \textbf{3.3\%} (82.1 to 85.4) and \textbf{3.7\%} (57.1 to 53.4) absolute gains on ICAcc and semER. In all cases, adding phone features during fine-tuning achieves a solid improvement on top of pre-trained models. One possibility is that the pre-training models are designed towards learning generic joint textual-phonetic representations; whereas the proposed fine-tuning approach exploits phone features directly for SLU.  Since computing WSA loss requires both $W$ and $P$ sequences, computing \emph{condMLM 100\% + condMSM 100\% (oneMod)} and WSA losses on the same training samples is infeasible. We add WSA to \emph{condMLM 30\% + condMSM 30\% (twoMod)} but it does not produce gain, probably because learning word and phone sequence alignment is relatively easy and a more difficult WSA task is required. We also evaluate using concatenation of $W$ and $P$  during fine-tuning (i.e., same as pre-training) instead of adding phone embedding, but did not observe consistent improvement over adding phone embedding. 

Our proposed pre-train and fine-tune approaches are agnostic to the ASR systems in cascaded SLU. All of our SLU fine-tuning use the SLU train set manual transcripts and NLU labels, and the same fine-tuned models are used for inferring intents and slots on ASR 1-best of valid and test sets generated by different ASR systems. In contrast, both prior works of improving SLU robustness to ASR errors~\cite{DBLP:conf/icassp/HuangC20,DBLP:conf/acl/HuangC20} require training on the specific ASR system.  \cite{DBLP:conf/icassp/HuangC20} finetunes ELMo using combined language model loss and confusion-aware loss on the ASR 1-best and WCNs of the SLU train set. The finetuned ELMo embeddings are used by biLSTM for SLU and achieved 89.55\% ICAcc on the Snips test set ASR output.
The two-stage SLU approach~\cite{DBLP:conf/acl/HuangC20} first pre-trains biLSTM on general domain text and then further pre-trains the model on ASR lattices of the SLU train set. Their approach achieves 95.37\% ICAcc on the Snips test set ASR hypotheses. Both prior approaches learn a significant amount of information about the specific ASR system, which may contribute to the substantial boost of SLU performance. We confirm this hypothesis by fine-tuning BERT-Base on the SLU train set ASR 1-best with NLU labels and evaluating ICAcc on the Snips test set ASR 1-best. This \textbf{Oracle} setup achieves 93.9 ICAcc w/o phone embeddings and 94.3 with PE. 
\newline
\noindent\textbf{Model Analysis.} To analyze the efficacy of the proposed pre-training model on aligning textual and phonetic representations, we select the most frequent 20 confusion word pairs from the FSC valid ASR 1-best (WER 36.7\%) and exclude pairs containing words not covered by the BERT vocabulary. Each word pair is used for retrieval with the ASR hypothesized word as query and the reference word as reference. We evaluate mean reciprocal rank (MRR) for retrieval by computing the cosine distance between the trained input embeddings from the pre-trained models. Table~\ref{tab:MRR-results} shows that compared to MRR 0.1012 from BERT-Base, the proposed pre-training model by further pre-training BERT-Base with condMLM 100\%+condMSM 100\% (oneMod) significantly improves MRR to 0.1591, confirming that the proposed pre-training model significantly reduces representation distance between acoustically confusable words.

\section{Conclusion}
\label{sec:conclusion}
We propose a novel pre-training approach to learn joint textual-phonetic representations for SLU. We design and study different pre-training tasks. We also propose incorporating phonetic features in fine-tuning. On FSC and Snips benchmarks, both proposed pre-training and fine-tuning approaches consistently improve SLU on ASR 1-best with different WERs and the gains are additive, achieving overall 2.7\% to 3.3\% absolute gain on intent accuracy and 3.7\% absolute gain on overall semantic frame accuracy, over strong baselines. Future work includes exploring more effective speech features and pre-training tasks.
\pagebreak

\bibliographystyle{IEEEtran}

\bibliography{refs}

\end{document}